%% file: anonymous-submission-latex-2024.tex
\documentclass[letterpaper]{article} 
\usepackage[]{aaai24}  
\usepackage{times}  
\usepackage{helvet}  
\usepackage{courier}  
\usepackage[hyphens]{url}  
\usepackage{graphicx} 
\usepackage{natbib}  
\usepackage{caption} 
\usepackage{algorithm}
\usepackage{algorithmic}

\usepackage{newfloat}
\usepackage{listings}
\DeclareCaptionStyle{ruled}{labelfont=normalfont,labelsep=colon,strut=off} 
\floatstyle{ruled}
\newfloat{listing}{tb}{lst}{}
\floatname{listing}{Listing}
\usepackage{amsmath}
\usepackage{amssymb}
\usepackage{mathtools}
\usepackage{amsthm}

\usepackage{subfigure}
\usepackage{multirow}
\usepackage{bm}
\usepackage{booktabs}
\usepackage{makecell}

\newcommand{\error}{element-wise error}

\newcommand{\verror}{dot-product error}
\usepackage[capitalize,noabbrev]{cleveref}

\usepackage{fixme}

\title{Efficient Adaptive Activation Rounding for Post-Training Quantization}
\author {
    Zhengyi Li\textsuperscript{\rm 1},
    Cong Guo\textsuperscript{\rm 1},
    Zhanda Zhu\textsuperscript{\rm 1},
    Yangjie Zhou\textsuperscript{\rm 1},
    Yuxian Qiu\textsuperscript{\rm 1},
    Xiaotian Gao\textsuperscript{\rm 3},
    Jingwen Leng\textsuperscript{\rm 1},
    Minyi Guo\textsuperscript{\rm 1},
}
\affiliations {
    \textsuperscript{\rm 1}Shanghai Jiao Tong University\\
    \textsuperscript{\rm 2}Microsoft\\
    \{hobbit,guocong,yj\_zhou,qiuyuxian,leng-jw\}@sjtu.edu.cn, zhandazhu@gmail.com, xiaotian.gao@microsoft.com, guo-my@cs.sjtu.edu.cn
}

\usepackage{bibentry}

\begin{document}

\maketitle

\input{tex_files/Abstract}

\input{tex_files/Introduction}

\input{tex_files/Preliminaries}

\input{tex_files/Motivations}
\input{tex_files/Methods}

\input{tex_files/Experiments}

\input{tex_files/RelatedWork}
\input{tex_files/Conclusion}

\bibliography{aaai24}

\end{document}


\begin{frontmatter}

\title{Appendix}



\end{frontmatter}

\input{tex_files/Appendix}


\bibliography{aaai24}

%% file: tex_files/Abstract.tex
\begin{abstract}
Post-training quantization attracts increasing attention due to its convenience in deploying quantized neural networks. Although rounding-to-nearest remains the prevailing method for DNN quantization, prior research has demonstrated its suboptimal nature when applied to weight quantization. They propose optimizing weight rounding schemes by leveraging output error rather than the traditional weight quantization error. Our study reveals that similar rounding challenges also extend to activation quantization. Despite the easy generalization, the challenges lie in the dynamic nature of activation. Adaptive rounding is expected for varying activations and the method is subjected to runtime overhead. To tackle this, we propose the AQuant quantization framework with a novel perspective to reduce output error by adjusting rounding schemes of activations. Instead of using the constant rounding border 0.5 of the rounding-to-nearest operation, we make the border become a function w.r.t. the activation value to change the activation rounding by the adaptive border. To deal with the runtime overhead, we use a coarse-grained version of the border function. Finally, we introduce our framework to optimize the border function. Extensive experiments show that AQuant achieves notable improvements compared to state-of-the-art works and pushes the accuracy of ResNet-18 up to 60.31\% under the 2-bit weight and activation quantization.
\end{abstract}


%% file: tex_files/Introduction.tex
\section{Introduction}
\label{sec:introduction}
Deep neural networks (DNNs) exhibit remarkable accuracy across diverse tasks. DNN quantization presents a solution by reducing both computational and storage burdens by converting floating-point numbers into low-bit representations. This technique has proven highly effective for deploying DNNs on devices with constrained resources. The quantization of DNNs can be categorized into two principal approaches. Quantization-aware training (QAT)\cite{esser2019learned, wang2019learning, gong2019differentiable} involves comprehensive fine-tuning of a pre-trained model using all available training data and extensive computational resources. However, this process is time-consuming and resource-intensive. In contrast, post-training quantization (PTQ)\cite{wang2020towards, hubara2021accurate, guo2021squant, NEURIPS2021_7cc23420, NEURIPS2021_ec895663, nagel2020up, li2020brecq, wei2022qdrop,frantar2023optq,xiao2023smoothquant} has garnered increasing attention due to its efficiency and ease of implementation. PTQ facilitates quantizing a pre-trained network using limited data (typically around 1K samples) and computational resources. Importantly, PTQ can effectively quantize model weights to 8-bit or 4-bit integers with negligible loss.


The operation of rounding-to-nearest is commonplace in both weights and activations quantization, aiming to minimize per-value rounding errors. 
However, recent works~\cite{nagel2020up, li2020brecq, wei2022qdrop} have unveiled that rounding weights to the nearest integers is suboptimal for the final output error, also referred to as the {\verror}. 
In contrast, they propose learning weight rounding schemes using output error as the optimization objective. 
This novel rounding scheme flips to a direction opposite the original, such as rounding 3.2 up to 4.0 rather than to the nearest 3.0. 
Although this may introduce greater errors for each element, it can ultimately reduce the final output error, allowing for quantization to 4 bits or even lower.
Notably, while they adopt a new rounding scheme for weight quantization, these methods continue to employ the rounding-to-nearest quantization for activations.

\begin{figure}[tbp] 
  \centering
    \includegraphics[width=0.98\linewidth]{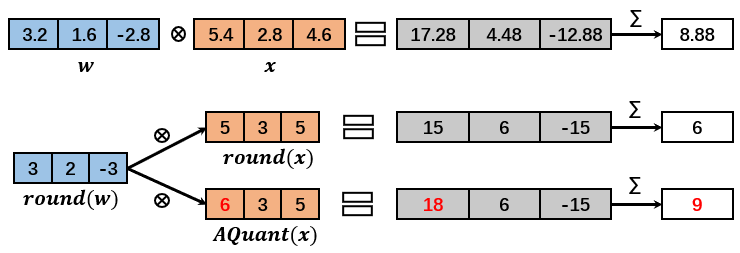}

  \caption{Illustration of adjusting activation rounding direction. Change the rounding of 5.4 from 5 to 6 effectively reduces the {\verror}.}
  \label{fig:illustration_x_flipping}
\end{figure}

This study observes that employing rounding-to-nearest for activations is also suboptimal. This phenomenon is visually depicted in Figure~\ref{fig:illustration_x_flipping}. 
While we retain rounding-to-nearest for weights to simplify the presentation, the quantized dot product ($6$) leads to a large error ($-2.88$) compared to the full-precision result ($8.88$).
Interestingly, if we modify the rounding of the first element, 5.4, of $x$ from 5 to 6, a smaller {\verror} ($+0.12$) can be achieved.
We later show such adjusting the activation rounding direction is purposeful.


Motivated by this observation, we propose AQuant, a new perspective of PTQ by adaptively rounding activations. 
However, different from weights with fixed rounding schemes, the dynamic nature of activations poses challenges.
First, given a weight value, it is reckless to force all activations that multiplied to it to be rounded down or up. 
To understand how adjust activation rounding can benefit the output error, we dive into the error of the element-wise multiplication of the dot product error. We find the expected value of the element-wise multiplication error can be manipulated to reduce the dot product error by changing the rounding border. 
Prior works~\cite{nagel2020up, li2020brecq, wei2022qdrop} can also be explained by the rounding border. The traditional rounding-to-nearest operation uses a rounding border of 0.5, and learning weight rounding schemes can be viewed as learning another constant border offline. 
For activations, as illustrated in Figure \ref{fig:illustration_x_flipping}, using a 0.3 rounding border for 5.4 achieves the expected adjustment. Moreover, unlike offline-learned weight rounding schemes, a predefined rounding border may not be optimal for every activation, so an adaptive border related to weights and activations is desirable.
This raises the second challenge that adjusting activation rounding during inference requires an efficient solution in terms of computation and memory.

To this end, AQuant aims to adaptively adjust the activation rounding via rounding border within negligible overhead. 
We first show that generalizing the prior constant border to a function of the activation value benefits the output error. 
To deal with the overhead, AQuant employs a coarse-grained version of the border function to maintain negligible runtime overhead. 
Equipped with this border function, we establish a PTQ framework that optimizes it jointly with weight rounding. 
The border function provides a new perspective to minimize the output error, supplementing weight rounding optimization to enable a larger search space. 
Extensive experiments on popular CNN models and the Transformer model~\cite{vaswani2017attention} show that our method significantly outperforms the recent state-of-the-art PTQ baselines~\cite{nagel2020up,li2020brecq,wei2022qdrop} with negligible extra parameters and computation overhead. Notably, AQuant recovers 60.31\% ResNet-18 accuracy with 2-bit weights and activations.

The contributions of this work are as follows:
\begin{enumerate}
    \item We theoretically show adaptive activation rounding schemes via border function can reduce the output error, providing a new perspective for the PTQ.
    \item Next, we provide an efficient deployment of the border function, enabling adaptive rounding border with negligible memory and computation overhead. 
    \item We finally propose a framework to optimize the proposed border function. Extensive experiments on various CNN and the transformer models show that AQuant greatly outperforms state-of-the-art PTQ methods under all the precision settings.
\end{enumerate}

%% file: tex_files/Preliminaries.tex
\section{Preliminaries}
\label{sec:preliminaries}

\paragraph{Notations}
The constant and scalar are denoted by italic letters, such as $x_i$ and $s$. The quantized value is indicated by the hat, such as $\hat{x_i}$. For layer $\ell$, we denote $\mathbf{W}^{(\ell)}$ as the weight tensor. Its input tensor is denoted by $\mathbf{X}^{(\ell)}$ and $\mathbf{x}^{(\ell)}$ represents one vector. We denote 
$\mathop{\mathbb{E}}[\cdot]$
as the expectation operator of $\mathbf{x}$ on the dataset. 

For the convolutional layer, we denote the output channel, input channel, and kernel size as $o_c$, $i_c$, and $k$, respectively. The height and width of the input feature map are $h_i$ and $w_i$, and $h_o$ and $w_o$ are the height and width of the output. 
To unify the feedforward of the convolution layer and the fully connected layer (FC) using matrix multiplication $\mathbf{X}^{(\ell+1)}= \mathbf{W}^{(\ell)} \mathbf{X}^{(\ell) }$ in the following discussion, we explicitly convert the convolution filter from ($o_c$, $i_c, k, k$) to ($o_c$, $i_c\times k\times k$). The input of convolutional layer is arranged from $(i_c,h_i,w_i)$ to $(i_c\times k\times k,h_o \times w_o)$ using the \texttt{img2col} transformation~\cite{paszke2019pytorch}. The $h_o \times w_o$ is the number of sliding blocks, and each one contains $i_c\times k\times k$ elements. Our discussion is mainly based on the CNN model. The activation function (ReLU) and the bias of the linear layer are omitted for simplicity. 

\paragraph{DNN Quantization} 

The quantization and dequantization for the scalar $x$ can be described as $\hat{x} =s \cdot \lfloor \frac{x}{s} \rceil$, where $\lfloor \cdot \rceil$ is the rounding-to-nearest operation and $s$ is the step size between subsequent quantization levels. We omit the clipping operation for simplicity. The rounding error of $x$ is $\Delta x=\hat{x} -x$. 
This work uses an equivalent notation of $\lfloor \cdot \rceil$
\begin{equation}
    \hat{x} =s \cdot \lceil x/s-B \rceil,
    \label{equ:rounding}
\end{equation}
which explicitly includes the rounding border $B$. We define the \textbf{rounding border} as any value with fractional part smaller than $B$ is rounded down and vice versa.
Traditional rounding-to-nearest operation minimizes the square value (or absolute value) of the rounding error $\Delta x$. In this way, the border $B$ is a constant 0.5. This work consider the {\error} as the objective and obtains a border that is a function regarding to the activation value.


The most common operation of DNN is the dot product between weight and activation vectors, which is the sum of element-wise multiplications $\mathbf{w}^T\mathbf{x}=\sum_i w_i x_i$~\cite{horn_johnson_1985}.
With the quantization, the error of the dot product is the sum of element-wise multiplication errors. We define the \textbf{element-wise (multiplication) error} as
\begin{equation}
\begin{aligned}
error^{EW} & =(w+\Delta w)(x+\Delta x)-w x \\
& =(w+\Delta w) \Delta x+\Delta w x,
\end{aligned}
\label{equ:derive_scalar_error}
\end{equation}
where $w$ is one element of vector $\mathbf{w}$ and $x$ is the corresponding activation value.

\paragraph{Post Training Quantization}
PTQ is performed based on a pre-trained model using only a small calibration dataset.
Early PTQ methods employ the rounding-to-nearest operation for both weights and activations, which minimizes the mean square error (MSE) between quantized and original values. Recent studies~\cite{nagel2020up,li2020brecq,wei2022qdrop} show that weight rounding should be learned using the layer output as the target.

\begin{equation}
\underset{\Delta \mathbf{W}_{k,:}}{\arg \min } \mathbb{E}\left[\left( \mathbf{\hat{W}}_{k,:} \mathbf{\hat{x}}-\mathbf{W}_{k,:} \mathbf{x}\right)^2\right],
\label{equ:weight_layer_wise_objective}
\end{equation}
where $\Delta \mathbf{W}_{k,:}=\mathbf{\hat{W}}_{k,:}-\mathbf{W}_{k,:}$ and $\mathbf{\hat{W}}_{k,:}$ can be either rounded up or down. $\mathbf{\hat{x}}$ is obtained by nearest rounding. Equ. \eqref{equ:weight_layer_wise_objective} decomposes the optimization of each layer to sub-problems. Each sub-problem deals with the dot product of a single weight vector $\mathbf{W}_{k,:}$. Its rounding scheme $\Delta \mathbf{W}_{k,:}$ should minimize the expected vector {\verror}. Early works optimize weight rounding schemes without activation quantization~\cite{nagel2020up,li2020brecq}. Later work has shown that including activation quantization into the optimization leads to better results~\cite{wei2022qdrop}.

%% file: tex_files/Methods.tex
\section{Motivation}
\label{sec:adaptive_rounding_difficulty}
This section presents our motivation and intuitive demonstration of AQuant's superiority over the previous weight-only scheme regarding its impact on element-wise multiplication and dot-product errors.

\paragraph{Element-wise Error.}
The adjustment of activation rounding in Figure \ref{fig:illustration_x_flipping} effectively reduces the {\verror}. To understand how the activation rounding affects the {\verror}, we consider the \textbf{\error}, a bridge between the rounding error and the {\verror}\footnote{We assume the {\error}s are independent for simplicity so that their expectations can sum up. Their correlation further complicates the equation but does not affect the conclusion.}:
\begin{equation}
\mathbb{E}\left[ \mathbf{\hat{w}}^T \mathbf{\hat{x}}-\mathbf{w}^T \mathbf{x}\right]=\sum_{i}^{K} \mathbb{E}\left[error^{EW}_{i}\right],
\label{equ:expeced_error_accumulation}
\end{equation}
where $K$ is the length of the vectors $\mathbf{x}$ and $\mathbf{w}$.
An implicit assumption of DNN quantization is that {\error}s are uniform; therefore, the effect of {\error}s can ``cancel out'' each other, resulting in a near-zero {\verror} when $K$ is large enough.
Unfortunately, when the length of the dot product is limited, and the quantization bit is ultra-low (e.g., 4-bit or lower), that assumption cannot be ensured, leading to {\error}s cannot be accumulated to near-zero.
As a result, the quantized dot product shifts away from the full-precision result, and the expected {\verror} is biased in a certain direction~\cite{finkelstein2019fighting, nagel2019data}.

\begin{figure}[tbp] 
  \centering
  \subfigure[Prior works that changes weight rounding schemes.]{\label{fig:NR_wx} \includegraphics[width=0.48\linewidth]{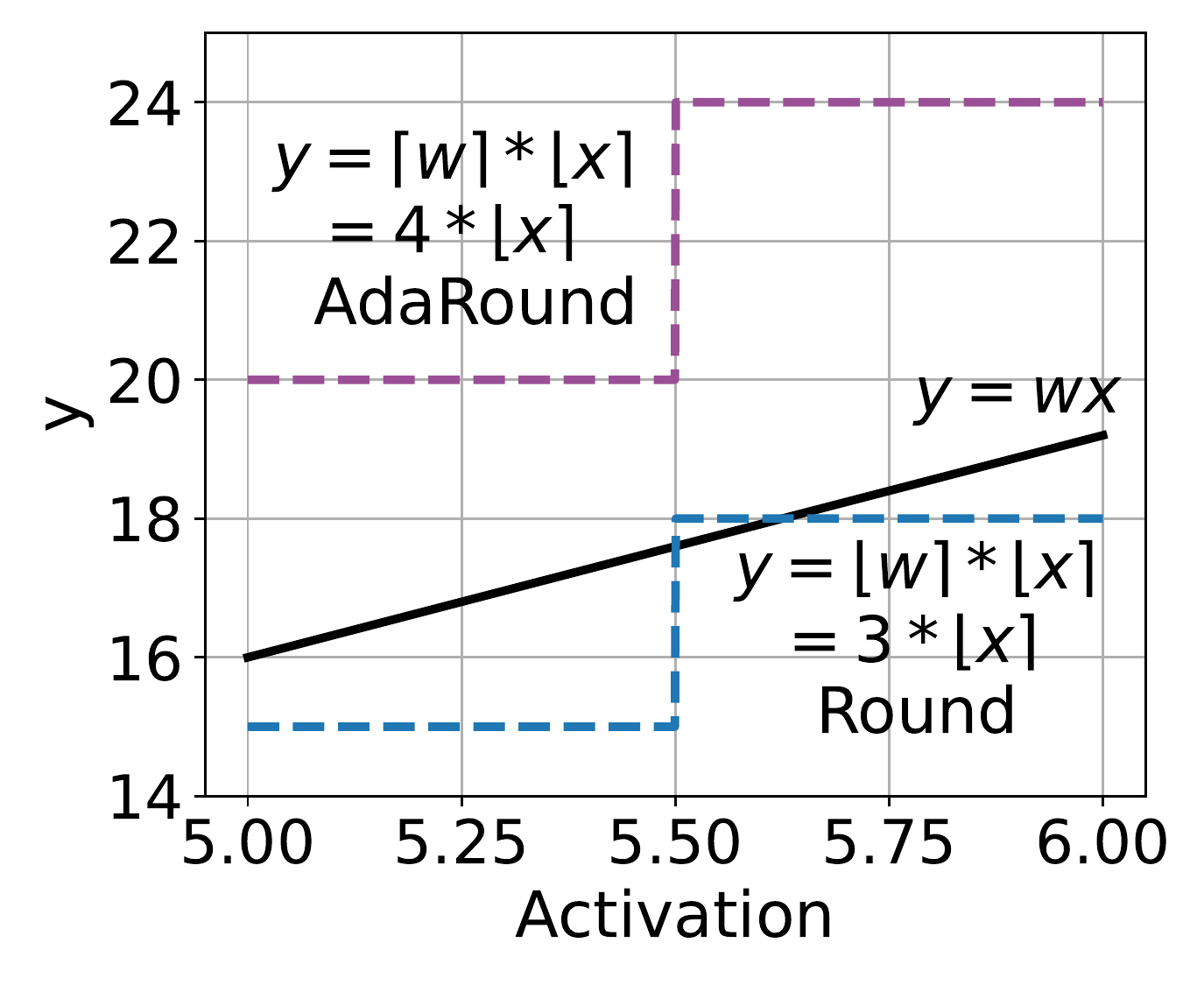}}
  \subfigure[Ours that changes the rounding border of activations.]{\label{fig:Flipped_wx}
  \includegraphics[width=0.48\linewidth]{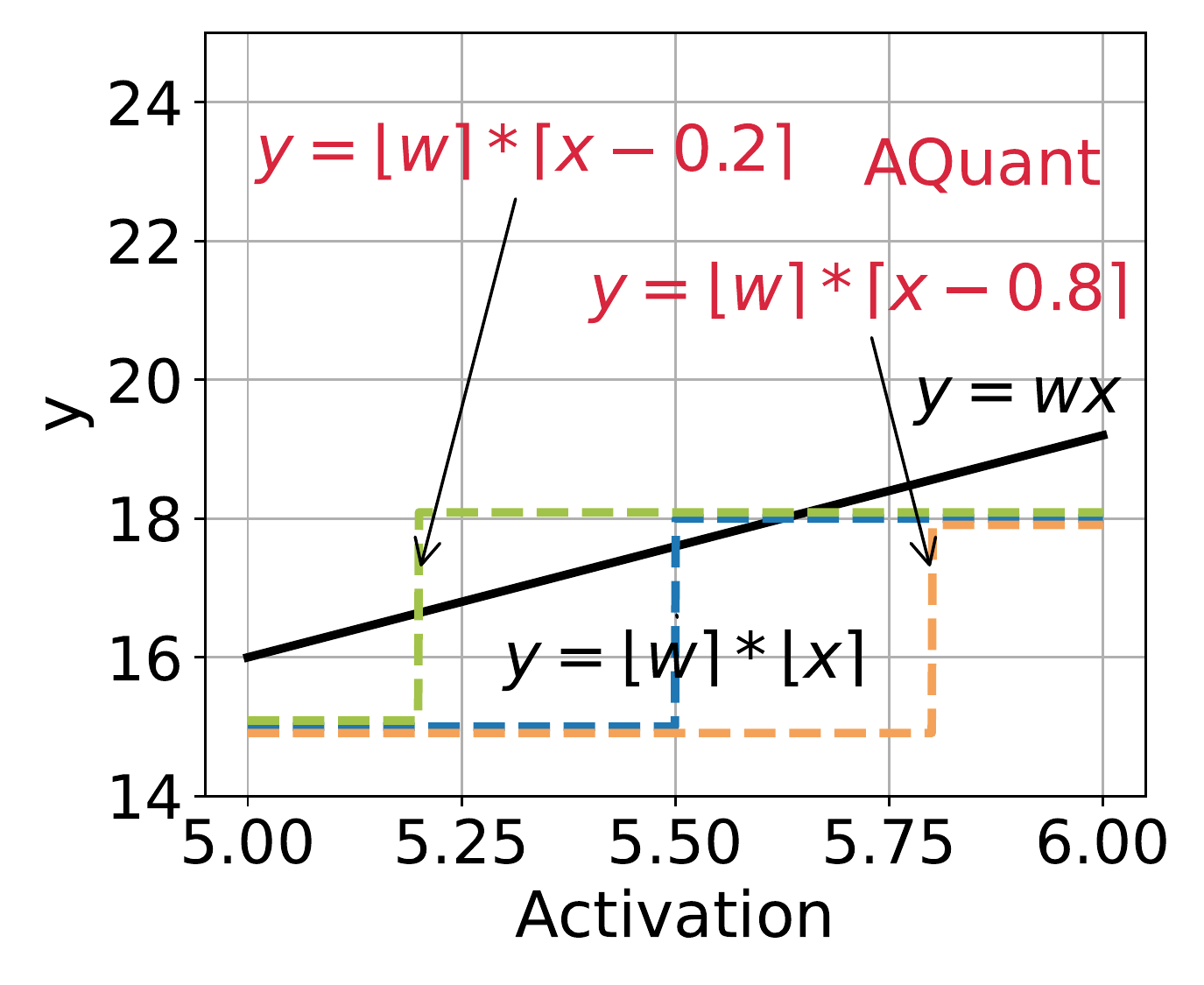}}
  \caption{Illustration of the manipulation of the expected value of the  {\error}, in which full-precision value $w=3.2$ and $x$ ranges from 5.0 to 6.0. }
  \label{fig:illustration_wx}
\end{figure}

\paragraph{AdaRound: Constrained Optimization Space.}
Many prior works have noticed such bias~\cite{finkelstein2019fighting, nagel2019data}. 
AdaRound-based works~\cite{nagel2020up,li2020brecq, wei2022qdrop} learn a new weight rounding scheme by Equ.~\eqref{equ:weight_layer_wise_objective} based on the ``flipping'' method, reducing the final {\verror}.
However, flipping the rounding direction of weight to the opposite direction always enlarges the expected {\error}.
To illustrate this issue, Figure~\ref{fig:NR_wx} shows a specific example ($w=3.2$) with $x$ ranging from 5.0 to 6.0.
The rounding results are close to the full-precision results (blue versus black curves), yielding negatively biased {\error}s.
But the AdaRound-based methods (purple curve) always deteriorate (enlarge) expected {\error} for the multiplication.
The flipping method enhances the cancellation among {\error}s, reducing the \verror. 
{However, this is constrained to flipping the rounding to the opposite direction and increases the expected element-wise multiplication error.}

\paragraph{AQuant: Flexible Optimization Space.}
Besides the AdaRound-based weight-only rounding, we can further adjust activation rounding to manipulate the {\error}.
For varying activations that are multiplied to the same weight value, it is reckless to assign all of them a fixed rounding direction. 
Therefore, we adjust the activation rounding through the rounding border, as defined in Sec. \ref{sec:preliminaries}. 
Changing the rounding border for activations can also manipulate the expected value of the {\error}. As depicted in Figure \ref{fig:Flipped_wx}, changing the border to 0.2 or 0.8 biases the expectation towards positive or negative, respectively (green and orange lines). 
Compared to flipping the weight rounding, which always increases the magnitude of the expected error, adjusting the activation rounding border also enables a reduction in the magnitude of the {\error} when manipulating its value, as exemplified by the green line in Figure \ref{fig:Flipped_wx}. Overall, the adjustment of the activation rounding border offers a new perspective to manipulate the expected {\error}, allowing for a broader exploration of potential solutions in a larger optimization space, which is the critical insight of AQuant. 




\section{AQuant}
Despite the effectiveness, adjusting activation rounding schemes poses two key challenges due to its dynamic nature.
First, unlike the weight rounding schemes fixed offline, activations are generated at runtime.  
Prior works that flip weight rounding can be viewed as offline learning a new constant rounding border. However, simply changing the rounding border of activations from 0.5 to another constant is insufficiently adaptive for diverse activations. It should be a function of weight and activation values. Second, the activation rounding is done during inference. The adjustment must be both computation- and memory-efficient. 
To deal with them, Sec. \ref{subsec:adaptive_rounding_border} first derives generalizing the constant rounding border to a border function related to the weight and activation value. Then Sec. \ref{subsec:practical_solution} provides a practical solution by using coarse-grained border function. Finally, Sec. \ref{subsec:aquant} details the optimization of the border function.

\subsection{Adaptive Rounding via Border Function}
\label{subsec:adaptive_rounding_border}
This section establishes a framework to understand the relationship between the weight, activation value and the {\error}. Using Equ. \eqref{equ:derive_scalar_error}, we consider the {\error} as an function of the activation rounding error. 
\begin{equation}
error^{EW}(\Delta x) =(w+\Delta w) \Delta x+\Delta w x,
\label{equ:scalar_error}
\end{equation}
where $w$ is one element of vector $\mathbf{w}$ and $x$ is the corresponding activation value.
We assume the weights and activations have been scaled by the step size in the following discussion. Both $w$ and $\Delta w$ are fixed at inference. Our discussion mainly focuses on the activation rounding error $\Delta x$.

As described in Sec. \ref{sec:preliminaries}, traditional rounding-to-nearest operation uses the self absolute difference as the objective, leading to a constant rounding border 0.5. To figure out how to adaptively adjust activation rounding for different activations, our core idea is to use the {\error} in Equ. \ref{equ:scalar_error} as the objective. Then the rounding border should depend on weight and activation values. Since the weight rounding is fixed offline, we write the border as a function w.r.t. the activation value $B^E(x)$. The superscript $E$ indicates the border is associated with one weight element. Similar to the rounding-to-nearest, if we let the {\error} of rounding up equals to down, it satisfies $|error^{EW}(-B^E(x))|=|error^{EW}(1-B^E(x))|$, from which we can derive
\begin{equation}
 B^E(x)= \frac{\Delta w}{w+\Delta w}x + \frac{1}{2}.
\label{equ:linear_border}
\end{equation}
Details of the derivation can be found in the Appendix. In this way, the quantized $\hat{x}$ is obtained by $\hat{x}=\lceil x-B^E(x) \rceil$. 
Compared to the constant border $B=0.5$ of the nearest rounding, $\frac{\Delta w x}{w+\Delta w}$ quantitatively defines the effects of the weight value, weight error, and activation value. For $\frac{\Delta w}{w+\Delta w}$, a larger magnitude of $\Delta w$ and a small magnitude of $w+\Delta w$ result in more border adjustments. Larger magnitudes of the activation value $x$ also lead to more border adjustments.

One interesting thing is that the border function of Equ. \eqref{equ:linear_border} can achieve the minimal and unbiased {\error}. First, we can show Equ. \eqref{equ:linear_border} achieves minimal {\error} by proving that any $x$ with a fractional part smaller than the rounding border has $|error^{EW}(\Delta x^-)|<|error^{EW}(\Delta x^+)|$ and vice versa, which we give formal proof in the Appendix. Second, the expected {\error} becomes unbiased. Assuming that the fractional part of $x$ is uniformly distributed in the interval $[0,1]$ for different inputs. Despite different integer parts of the $x$, the expected {\error} concerning various $x$ within an integer interval is the integral of Equation \eqref{equ:scalar_error} from $-B$ to $1-B$. 
\begin{equation}
 \mathbb{E}[error^{EW}] =\int_{-B}^{1-B} \left[(w+\Delta w) \Delta x+\Delta w x \right] d \Delta x
\end{equation}
In the nearest rounding, where $B=0.5$, the expected error value is $\Delta w x$. Even though previous studies have optimized the weight rounding schemes, the bias persists since $\Delta w$ always exists. In contrast, an adaptively adjusted rounding border $B^E(x)$ can make the expected {\error} become zero.

\begin{figure}[tbp] 
  \centering
  \subfigure[The optimal way to round activation using Equ. \eqref{equ:linear_border}.]{\label{fig:optimal_method} \includegraphics[width=0.47\linewidth]{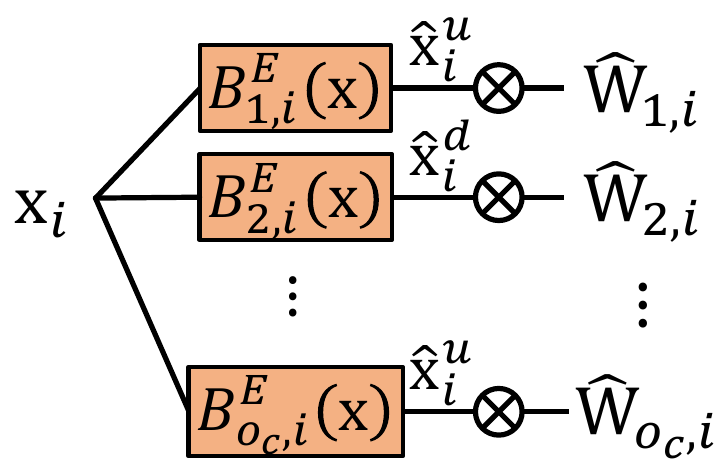}}
  \
  \
  \
  \subfigure[Coarse-grained border function. $o_c$ weights only need one border function.]{\label{fig:approximated_method}
  \includegraphics[width=0.47\linewidth]{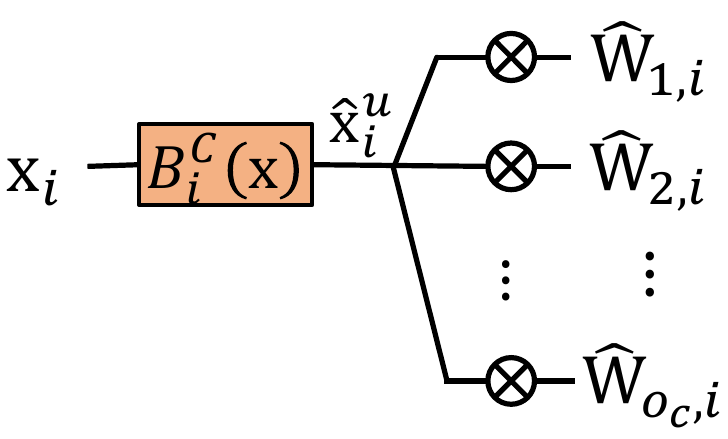}}
  \caption{Illustration of two types of activation rounding, using the $i_{th}$ column of weight tensor $\hat{W}_{:,i}$. The corresponding activation $x_i$ is the $i_{th}$ value of the activation vector. Rounding up or down is indicated by superscripts 'u' and 'd'.}
  \label{fig:illustration_method}
\end{figure}

\subsection{Towards Practical Solution} 
\label{subsec:practical_solution}
The next challenge lies in efficiently applying the border function at the inference stage. Despite the promising results of the border function Equ. \eqref{equ:linear_border}, directly deploying it is not feasible due to its overhead and incompatibility with existing GPU kernels. 
The border function is one-to-one associated with weight values. Each border function needs to store one coefficient so that there are as many extra parameters as the model weights, which is unacceptable. Similarly, for each $wx$, we need to compute $B^E(x)$ once, which requires equal number multiplication than the inherent number of the matrix multiplication.
Moreover, as shown in Fig. \ref{fig:optimal_method}, each activation value $x_i$ is multiplied with $o_c$ weight values in the matrix multiplication, where $o_c$ denotes the number of output channels. Rounding $x$ differently for each weight is incompatible with the existing GPU computation kernel of the matrix multiplication. 
This work aims to provide a feasible solution based on existing DNN frameworks. Co-designing software and hardware could potentially provide a better solution, and we leave it for future work.

Instead of using exact Equ. \eqref{equ:linear_border} for optimal {\error}, we can regard it as a tool to adjust the {\error}. Suppose one weight value is associated with a border function $B^E(x)=0.2x+0.5$, which makes the expected {\error} become zero. Assigning $B^E(x)=0.1x+0.5$ or $B^E(x)=0.3x+0.5$ can push the expected value positive or negative. 


Based on prior insights, we increase the application granularity of the border function. We let one column of weights share one border function. Each activation value is only rounded once, and its rounding scheme is shared by $o_c$ weights, as illustrated in Fig. \ref{fig:approximated_method}.
In this way, the expected value of the {\error} of the $o_c$ weights are either towards or away from zero. Similar to prior works that adjust the weight rounding, such adjustment provides a new perspective to manipulate the {\error} for better cancellation. Overall, we are more flexible in adjusting the {\error} than only learning the weight rounding schemes.
The coefficients of the coarse-grained border are hard to determine manually. Therefore, we parameterize the border function as 
\begin{equation}
B^{o_c}(x)=b_{1} \cdot x+b_{0},
\label{equ:parameterized_linear_border}
\end{equation}
and learn its parameters on the calibration set together with other learnable parameters. The superscript "$o_c$" indicates it is shared by weights from $o_c$ output channels.

\begin{figure}[tbp] 
  \centering
  \includegraphics[width=0.99\linewidth]{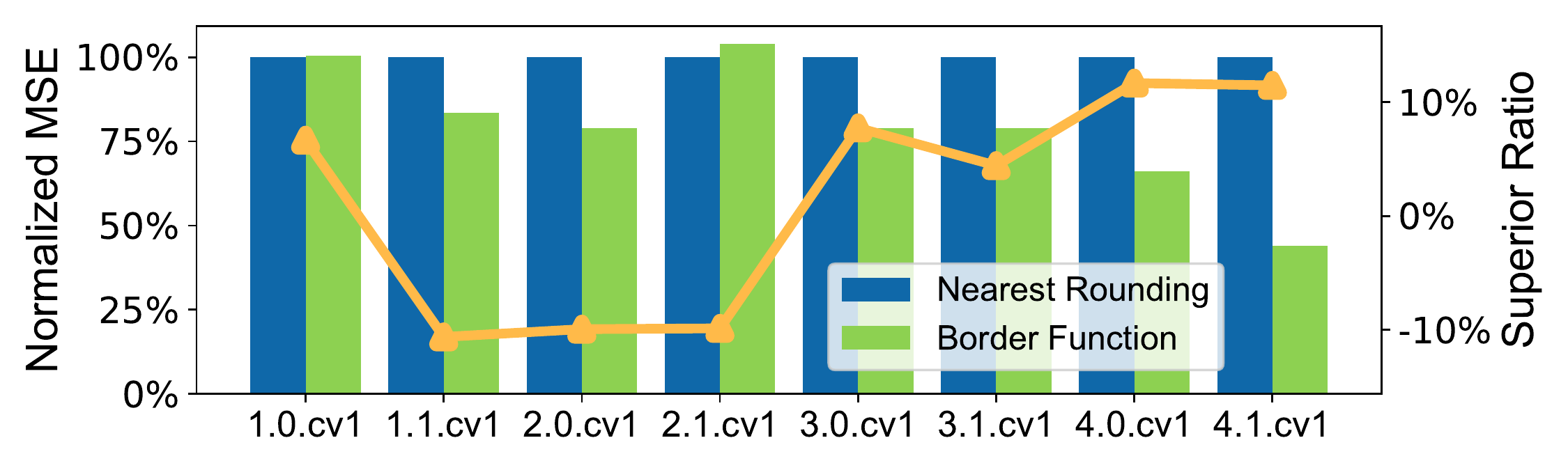}
  \caption{Effects of the approximated border function. The bars correspond to the left y-axis and are normalized based on the MSE of the nearest rounding. The curve is associated with the right y-axis.}
  \label{fig:bar_error_verification}
\end{figure}
The usage of Equ. \eqref{equ:parameterized_linear_border} offers more flexibility to adjust the {\error} than merely adjust weight rounding schemes. To validate the effectiveness of Equ. \eqref{equ:parameterized_linear_border}, we provide a concrete example using ResNet-18 under W2A4 quantization and 64 images. We compare our method with prior works that only learn the weight rounding schemes. Details of training the border function follow the setting in Sec. \ref{subsec:aquant}. We first present MSE errors from layers of varying depths in Fig. \ref{fig:bar_error_verification}. Using our border function achieves lower MSE on almost all layers towards the oracle full-precision activations. We find the enhancements are particularly pronounced in deeper layers, contributing to improved final accuracy. Then we delve deep into {\error}s. We call a rounding scheme of activation value superior if our border function results in a smaller {\error}  than the nearest rounding. The superior ratio indicates the ratio that the border function is superior across different positions of the feature map. As the curve shows, we achieve a higher ratio on most of the layers, especially in deeper ones. We also observe that the border function performs worse than nearest rounding for some layers. However, considering it attains smaller output MSE, we attribute this observation to its emphasis on error cancellation, which may result in more numbers of bad changes.

\paragraph{Overhead Analysis} 
We analyze the overhead using the convolutional layer and a similar analysis can also be applied to the fully connected layer. We first analyze the extra parameter ratio of the border function. For the reshaped convolutional filter with shape $(o_c, i_c\!\times\!k^2)$, since we let all output channels share the border function, we need $ i_c\!\times\!k^2$ border functions. The two parameters in the border function double it and result in $2\!\times\!i_c\!\times\!k^2$ extra parameters, which is $2/o_c$ of model weights and is very small due to the hundreds or thousands $o_c$. 
Then we analyze the computing overhead. Taking the scalar multiplication as the basic operation, the extra computation is only $O(1/o_c)$ compared to the innate convolution.
This ratio is also because we only compute the border once when multiplying with $o_c$ weights. Moreover, $B(x)$ only includes element-wise computation. Such a computation pattern is more efficient than the dominating operation of DNNs, matrix multiplication, in the sense of computing parallelism and memory access. The computation of the border function can be easily fused with other kernels~\cite{niu2021dnnfusion}, such as the \texttt{img2col} or previous \texttt{ReLU}. We implement the fused kernel and give experiments in Sec. \ref{subsec:overhead_analysis} to show the efficiency. 
In summary, the memory and computation only incur $O(1/o_c)$ extra overhead. Even though we assign the border function a higher bitwidth than the backbone model, the overhead is still small enough to ignore.

\subsection{Applying AQuant for Optimization}
\label{subsec:aquant}



With the effective and efficient border function in Equ. \eqref{equ:linear_border}, we integrate its optimization Equ. \eqref{equ:weight_layer_wise_objective}.
Substituting the $\Delta \mathbf{x}$ by $\lceil \mathbf{x}-B(\mathbf{x}) \rceil-\mathbf{x}$, the optimization objective becomes:
\begin{equation}
\begin{split}
 \min_{B^{o_c}(\mathbf{\cdot}), \Delta \mathbf{W}} \mathbb{E}\left[\left\Vert(\mathbf{W}+\Delta \mathbf{W}) (\lceil \mathbf{x}  -  B^{o_c}(\mathbf{x}) \rceil)  -\mathbf{W} \mathbf{x}\right\Vert_{2}^{2}\right].
\label{equ:final_aquant_goal}
\end{split}
\end{equation}

Compared to direct writing Equ. \eqref{equ:weight_layer_wise_objective}, We here include all weights of a layer in the objective. We suppose all values are scaled by scale steps for simplicity, but note they are also optimized together with other parameters and the used method follows LSQ~\cite{esser2019learned}. The application of $B^{o_c}(\mathbf{x})$ follows Figure \ref{fig:approximated_method}. The number of border functions equals the length of the activation vector, which is also the hidden dimension of the weight $\mathbf{W}$. Then each border function is applied to the corresponding entry of the activation vector. The optimization of weight quantization follows previous methods~\cite{nagel2020up,li2020brecq,wei2022qdrop}, and we only denote optimizing $\Delta \mathbf{W}$ for simplicity. We use gradient descent to optimize the above formula in layer fashion~\cite{nagel2020up} or block fashion~\cite{li2020brecq,wei2022qdrop}. To stabilize the optimization of the border function, we also design specific optimization techniques and we give details in the Appendix.


%% file: tex_files/Experiments.tex
\section{Experiments}
\label{sec:experiments}

\begin{table*}[tbp]
\footnotesize
  \centering
  \caption{Accuracy comparison on fully quantized models. All results include LSQ during optimization.}
    \begin{tabular}{cccccccc}
    \toprule
    Methods & Bits  & Res18 & Res50 & MNV2  & Reg600MF & Reg3.2GF & MNasx2 \\
    \midrule
    Baseline & W32A32 & 71.01 & 76.62 & 72.63 & 73.53 & 78.45 & 76.52 \\
    \midrule
    AdaRound & W4A4  & 69.49 & 74.82 & 67.53 & 70.87 & 76.41 & 71.80 \\
    BRECQ & W4A4  & 69.45 & 75.05 & 68.29 & 70.96 & 76.64 & 73.24 \\
    QDrop & W4A4  & 69.61 & 75.40 & 68.57 & 71.19 & 76.76 & 73.54 \\
    AQuant & W4A4  & \textbf{70.03} & \textbf{75.58} & \textbf{69.15} & \textbf{71.68} & \textbf{77.03} & \textbf{73.78} \\
    \midrule
    AdaRound & W2A4  & 63.64 & 66.61 & 44.90 & 58.95 & 64.66 & 51.03 \\
    BRECQ & W2A4  & 64.84 & 70.27 & 53.43 & 62.41 & 71.21 & 62.11 \\
    QDrop & W2A4  & 65.16 & 70.71 & 54.22 & 63.59 & 71.34 & 63.78 \\
    AQuant & W2A4  & \textbf{66.63} & \textbf{71.11} & \textbf{56.74} & \textbf{65.71} & \textbf{72.98} & \textbf{65.27} \\
    \midrule
    AdaRound & W3A3  & 66.00 & 69.72 & 48.92 & 61.99 & 68.65 & 50.48 \\
    BRECQ & W3A3  & 66.01 & 71.73 & 54.66 & 63.79 & 71.75 & 64.16 \\
    QDrop & W3A3  & 66.65 & 72.38 & 57.68 & 65.37 & 72.69 & 66.43 \\
    AQuant & W3A3  & \textbf{67.97} & \textbf{73.20} & \textbf{58.56} & \textbf{67.11} & \textbf{74.09} & \textbf{68.96} \\
    \midrule
    BRECQ & W2A2  & 51.12 & 51.36 & 7.44  & 29.84 & 41.88 & 13.32 \\
    QDrop & W2A2  & 54.19 & 57.38 & 11.45 & 41.53 & 54.61 & 28.65 \\
    AQuant & W2A2  & \textbf{60.31} & \textbf{60.04} & \textbf{14.19} & \textbf{46.61} & \textbf{57.48} & \textbf{30.31} \\
    \bottomrule
    \end{tabular}%
  \label{tab:fully}%
\end{table*}%

\begin{table*}[htb]
\footnotesize
  \centering
  \caption{W4A6 quantization results on GELU benchmark (except WNLI). All cases include LSQ during optimization. $^{\dagger}$Reports F1 score for MRPC and QQP. $^{\ddagger}$Reports Pearson correlation for STS-B. $^{\psi}$Reports matched accuracy for MNLI.}
    \begin{tabular}{ccccccccc}
    \toprule
    \multicolumn{1}{l}{Method} & CoLA  & SST-2 & MRPC$^{\dagger}$ & STS-B$^{\ddagger}$ & QQP$^{\dagger}$ & MNLI$^{\psi}$ & QNLI  & RTE \\
    \midrule
    FP baseline & 58.59  & 91.28  & 89.54  & 89.14  & 87.98  & 84.06  & 91.19  & 70.39  \\
    \midrule
    AdaRound & 43.82  & 82.45  & 87.92  & 83.13  & 74.24  & 65.77  & 83.29  & 55.60  \\
    AQuant & \textbf{45.23} & \textbf{84.06} & \textbf{88.62} & \textbf{83.87} & \textbf{75.87} & \textbf{68.57} & \textbf{83.45} & \textbf{61.37} \\
    \bottomrule
    \end{tabular}%

  \label{tab:GELU_results}%
\end{table*}%

To verify the effectiveness of AQuant, we first compare AQuant with other SOTA PTQ approaches. Then we demonstrate the negligible overhead of AQuant. 

\paragraph{Experimental Setup} Our algorithm is implemented using PyTorch~\cite{paszke2019pytorch} and evaluated on Nvidia GPU V100. The experiments include image classification on ImageNet and natural language processing tasks on the GLUE benchmark. On both of them, we sample 1024 instances as the calibration set, as the baseline works do. The weight quantization follows AdaRound on GLUE benchmark~\cite{nagel2020up} and QDrop~\cite{wei2022qdrop} on the ImageNet dataset and adopts per-channel weight quantization. The step size of activation quantization follows LSQ~\cite{esser2019learned}. Prior works~\cite{nagel2019data,peg} learn the weight rounding without activation quantization. Then they fix the weight quantization and learn the activation quantization scale step. Later works~\cite{wei2022qdrop} quantize both weights and activations and jointly learn optimization, which produces better optimization results. For fairness, we jointly optimize all parameters for AQuant and all baselines. More detailed settings are described in the Appendix.


\paragraph{Baselines} As far as we know, we are the first work that optimizes the activation rounding. We choose works that are closest related to ours. Prior works, AdaRound~\cite{nagel2020up}, BRECQ~\cite{li2020brecq}, and QDrop~\cite{wei2022qdrop} target the output error via learning the weight rounding. AdaRound and BRECQ optimize the weight rounding schemes layer-wisely and block-wisely, respectively. QDrop further randomly drops the quantization of activations when optimizing weight rounding. For works about activation quantization, previous works mainly optimize the step size (or clipping range) and we pick LSQ~\cite{esser2019learned} as the used method. We do not list it separately but integrate it in all experiments since merely using it suffers great accuracy loss for low-bit quantization.

\subsection{Comparison with SOTA Methods }
\label{subsec:comparision_exp}
\paragraph{ImageNet} To evaluate the effectiveness of AQuant, we conduct experiments on 6 CNN models using ImageNet~\cite{5206848} dataset, including ResNet-18 \& 50~\cite{he2016deep}, MobileNetV2~\cite{sandler2018mobilenetv2}, MNasNet~\cite{tan2019mnasnet} and RegNet~\cite{radosavovic2020designing}. The models we use cover common designs of CNNs, including the residue connection, depthwise-separable convolution, and group convolution. 
Results of all baselines and our implementation are based on the open-source codes of QDrop~\cite{wei2022qdrop} with unified settings.

We present the results of fully-quantized models in Table \ref{tab:fully}. We focus on quantization lower than 8-bit as W8A8 is almost loss-free in previous works. Our method outperforms previous PTQ methods in all cases.
The benefits of AQuant become more prominent as the bitwidth decreases. For challenging models like MobileNetV2 and MNasNet, we also achieve significant improvements compared with other methods. We find that ResNet-18 and ResNet-50 achieve more than 60\% accuracy in the extremely low W2A2 quantization. It is also worth noting that AQuant only has around 1\% accuracy loss for 4-bit quantization on many models. 

\paragraph{GLUE Benchmark} Although our method is discussed mainly on the convolutional layer, it can be applied to the FC layer of the transformer model. We use the fine-tuned BERT model and the GLUE benchmark from Huggingface~\cite{huggingface}. Our implementation is based on the open-sourced code of \cite{peg}, which implements AdaRound. We do not compare with QDrop since they do not release their codes on the BERT. Our method is orthogonal to them and the comparison with AdaRound is sufficient to demonstrate the effectiveness of AQuant. All the intermediate activations are quantized except the embedding and the classification layers. Other settings are kept the same. We adopt W4A6 to quantize the transformer model. AQuant surpasses the baseline in all tasks. 
Our method averagely improves the accuracy by nearly two percent, even when the results are already close to full-precision accuracy. 


\subsection{Overhead Analysis}
\label{subsec:overhead_analysis}
\paragraph{Ratio of AQuant Parameters} AQuant introduces only $O(1/o_c)$ extra number of parameters. Although existing works have shown that 8-bit quantization can achieve almost lossless accuracy for DNNs~\cite{jacob2018quantization}, we post-quantize the border function to 12 bits to guarantee precision. For ResNet-18 and ResNet-50, ratios of the extra number of parameters are 0.81\% and 0.64\%, respectively. If weights utilize 4-bit quantization, AQuant only increases 2\% of the model size.  For RegNet-600MF and RegNet-3200MF, the ratios of the extra number of parameters are 2.82\% and 2.14\%, and the extra model size is less than 8\%. For small models MobileNetV2 and MNasNet, the ratios of the extra number of parameters are 4.56\% and 8.27\%, and the extra model size is around 24\%. As for the transformer model, the output channel is at least 768. The extra model size is only 0.75\%. We can further reduce the extra size by directly training quantized border functions, such as 8-bit quantization. 

\begin{figure}[tbp] 
    \centering 
    \includegraphics[width=0.85\linewidth]{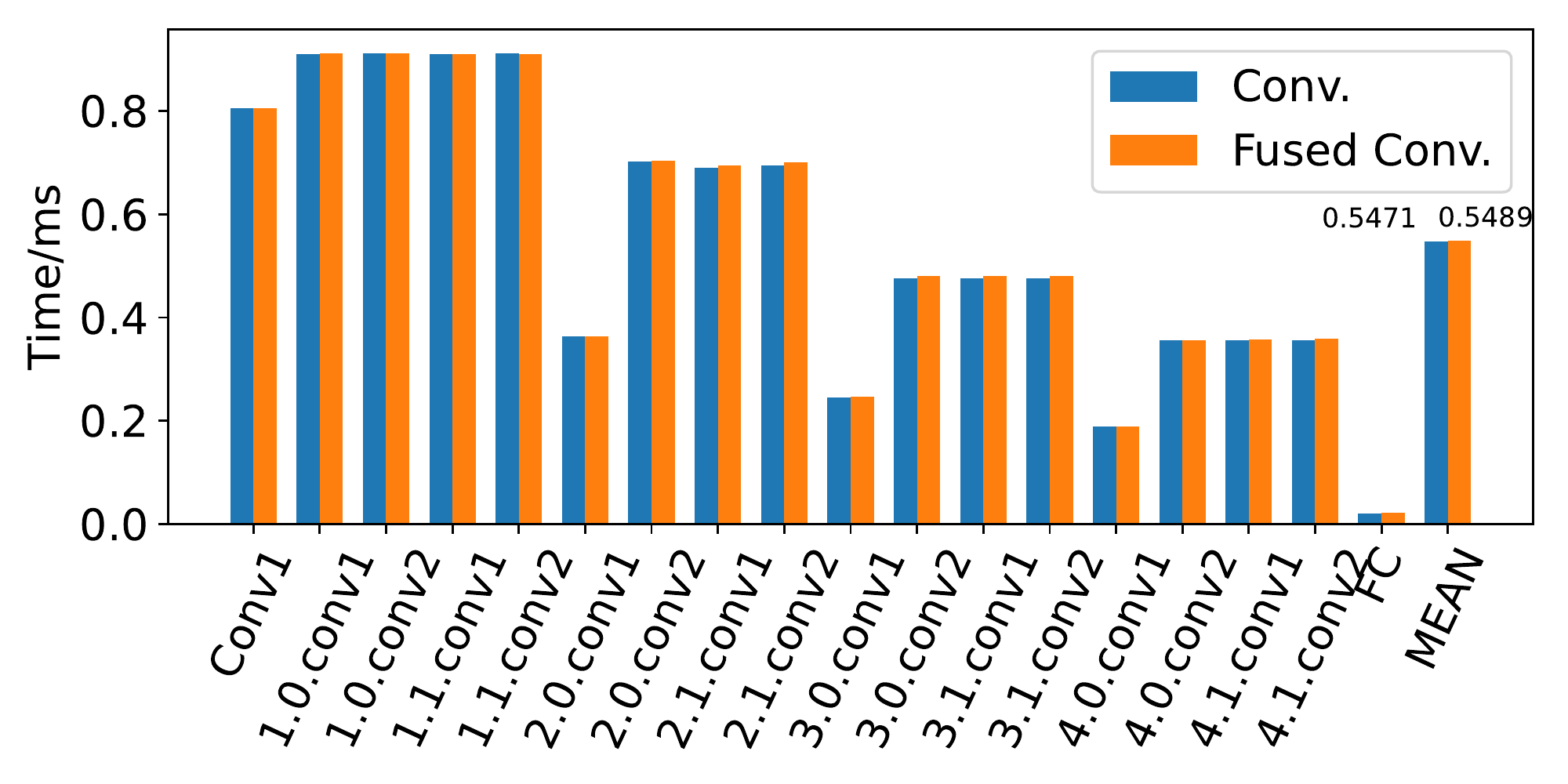}  
    \caption{Latency breakdown on the ResNet-18. The experiment is executed 100 times and reports the average value.} 
    \label{fig:latency_breakdown}
\end{figure}

\paragraph{Runtime Overhead} AQuant introduces merely $O(1/o_c)$ additional computational complexity. Moreover, the overhead of AQuant is even more negligible due to the benefits of kernel fusion. It reduces the overhead associated with kernel launching and memory swapping, which have been shown to be more dominant than the computation~\cite{niu2021dnnfusion}. Owing to the element-wise computation pattern, the border function can be easily fused with other computation kernels, such as \texttt{ReLU} and \texttt{img2col}. We choose to fuse it with the \texttt{img2col}. Our implementation is based on the open-source code of Caffe~\cite{jia2014caffe} and is tested on the Nvidia V100. Although we cannot run real low-bit inference, similar results should be expected with other DNN frameworks or hardware. Utilizing a batch size of 32, we list the execution times for all layers of ResNet-18. As demonstrated in Fig. \ref{fig:latency_breakdown}, the execution times of the fused convolution are close to the original times. For the average layer latency, the additional latency is 0.0028ms (the MEAN bar), constituting only 5.11\textperthousand~of the original inference time. Note that we only implement a naive fused kernel, and a more optimized implementation could yield better performance.
In practice, the border function requires a higher bitwidth than the innate computation of neural network models. For example, we may use W12A12 to compute the border for a W3A3 quantized model. Theoretically, the higher bitwidths incur $4 * 4=16$ times slower latency, which is $16*$5.11\textperthousand$=8.17\%$. But the extra latency ratio is still small enough to ignore.

%% file: tex_files/RelatedWork.tex
\section{Related Work}
\label{sec:related_work}
Quantization of deep neural networks can be divided into two categories: Quantization-Aware Training (QAT) and Post-Training Quantization (PTQ). QAT trains a quantized model from scratch using the full training dataset, while PTQ fine-tunes a pre-trained model with a limited dataset (typically 1k images). Although QAT has achieved promising accuracy levels~\cite{esser2019learned, wang2019learning, gong2019differentiable}, it is burdened by high training costs and hyperparameter search. In contrast, PTQ has attracted considerable interest due to its convenience. Studies in this area have focused on optimizing the weight clipping range~\cite{choukroun2019low,banner2019post}, splitting channels to exclude outlier values for lower bit quantization at the expense of additional computation~\cite{zhao2019improving}, mixed precision~\cite{cai2020zeroq}, and data-free quantization~\cite{guo2021squant}.

The most relevant track is optimizing the weight rounding schemes. AdaRound~\cite{nagel2020up} first provides a theoretical demonstration that nearest rounding, which minimizes the MSE of the weight itself, is suboptimal. Instead, they the weight rounding should serve for the output MSE and learn it in a layer-wise fashion. BRECQ~\cite{li2020brecq} further introduces block-wise weight scheme learning. QDrop~\cite{wei2022qdrop} drops activation quantization during learning to import flatness. However, we demonstrate that optimized weight rounding schemes still result in biased and magnified {\error}s. These methods are also not applicable to activation rounding due to dynamicity. Prior research about activation quantization tends to neglect activation rounding because of the difficulties associated with its dynamicity and runtime overhead and instead focuses on optimizing activation quantization from other perspectives. Some studies~\cite{banner2018aciq, choi2018pact, esser2019learned} optimize the step size to minimize the MSE of the activations themselves. Others~\cite{DBLP:journals/corr/abs-1906-03193,nagel2019data} compensate for the mean activation shift by rectifying the linear layer's bias. However, all these works assume a rounding-to-nearest operation at the rounding stage. In this research, we adaptively round the activations using a lightweight border function, presenting a novel perspective for PTQ.

%% file: tex_files/Conclusion.tex
\section{Conclusion}
\label{sec:conclusion}
This paper presents AQuant, a novel approach that makes up for the blank of activation rounding in PTQ algorithms. We first show that an adaptive rounding border can manipulate the {\error}, which further reduces the {\verror}. Based on this insight, we use a coarse-grained border function to round activations at runtime efficiently. AQuant improves accuracy across all bitwidths with significant improvements in low bitwidths, while introducing negligible extra parameters and computational overhead. Future work includes leveraging this insight to develop more effective border functions and reduce runtime overhead through algorithm and hardware co-design.

%% file: tex_files/Appendix.tex
\renewcommand\thesection{\Alph{section}}

\appendix
The following discussions assume the values are scaled by the step size and we follow notations in the paper. We first give the derivation of the border function in Sec. \ref{appendix:derivation}. Sec. \ref{appendix:implementation} introduces designs of AQuant aiming at optimizing the border function. Sec. \ref{appendix:experimental_setup} mentions detailed setups of our experiments.

\section{Derivation of the Border Function}
\label{appendix:derivation}
\paragraph{Linear border}
Recall the {\error} $\left[(w+\Delta w) \Delta x+\Delta w x\right]$ in Sec. 2, we give the derivation of Equ. (6) of the original paper. For the rounding border, the square (or absolute value) of the error function for rounding up and rounding down should have the same value $error^{EW}(-B^E(x))^2=error^{EW}(1-B^E(x))^2$. 
\begin{equation}
\footnotesize
\begin{aligned}
error^{EW}(-B^E(x))^2= &error^{EW}(1-B^E(x))^2 \\
(w+\Delta w)(-B^E(x))+\Delta w x= &-((w+\Delta w)(1-B^E(x))+\Delta w x) \\
(w+\Delta w)(2 B^E(x)-1)= &2 \Delta w x \\
 B^E(x)= &\frac{\Delta w}{w+\Delta w}x + \frac{1}{2}.\\
\label{equ:derivation_of_linear_border}
\end{aligned}
\end{equation}


\section{Proof of the Border Property}
\label{appendix:border_property}
Recall in the rounding-to-nearest operation, any numbers with fractional part smaller than 0.5 are rounded down and greater than 0.5 are rounded up, which minimizes the rounding error itself.
For the border function in Equ. (6) of the original paper, we can also show it achieve minimal {\error} by verifying the following theorem.
\begin{theorem}
A border $B$ is valid if $\forall \text{ x with fractional part } frac(x)<B^E(x), error^{EW}(\Delta x^{-})^2<error^{EW}(\Delta x^{+})^2$, and vice versa.
\end{theorem}

\begin{proof}

We first give the proof of the case when $frac(x)<\frac{\Delta w}{w+\Delta w} x+\frac{1}{2}$. By substituting the $\Delta x^{-}=-frac(x)$ and $\Delta x^{+}=1-frac(x)$, we have
\begin{equation}
    \begin{aligned}
    error^{EW}\left(\Delta x^{-}\right)^2 & =\left[(w+\Delta w) \Delta x^{-}+\Delta w x\right]^2 \\
    & =[(w+\Delta w)(-\operatorname{frac}(\mathrm{x}))+\Delta w x]^2 \\
    & =[(w+\Delta w) \operatorname{frac}(\mathrm{x})]^2-2 \Delta w x(w+\Delta w) \operatorname{frac}(x)+(\Delta w x)^2,
    \end{aligned}
\end{equation}
and
\begin{equation}
    \begin{aligned}
    error^{EW}\left(\Delta x^{+}\right)^2= & {\left[(w+\Delta w) \Delta x^{+}+\Delta w x\right]^2 } \\
    = & {[(w+\Delta w)(1-\operatorname{frac}(\mathrm{x}))+\Delta w x]^2 } \\
    = & {[(w+\Delta w)(1-\operatorname{frac}(\mathrm{x}))]^2+2 \Delta w x(w+\Delta w)(1-\operatorname{frac}(x))+(\Delta w x)^2 } \\
    = & {[(w+\Delta w) \operatorname{frac}(\mathrm{x})]^2-2(w+\Delta w)^2 \operatorname{frac}(x)+(w+\Delta w)^2-} \\
    & 2 \Delta w x(w+\Delta w)(\operatorname{frac}(x))+2 \Delta w x(w+\Delta w)+(\Delta w x)^2.
    \end{aligned}
\end{equation}

Subtracting $error^{EW}\left(\Delta x^{-}\right)^2$ by $error^{EW}\left(\Delta x^{+}\right)^2$ and import $frac(x)<\frac{\Delta w}{w+\Delta w} x+\frac{1}{2}$, we have
\begin{equation}
    \begin{aligned}
    & error^{EW}\left(\Delta x^{-}\right)^2-error^{EW}\left(\Delta x^{+}\right)^2 \\
    =& -2(w+\Delta w)^2 \operatorname{frac}(x)+(w+\Delta w)^2+2 \Delta w x(w+\Delta w) \\
    <& -2(w+\Delta w)^2\left(\frac{\Delta w}{w+\Delta w} x+\frac{1}{2}\right)+(w+\Delta w)^2+2 \Delta w x(w+\Delta w) \\
    =& -2 \Delta w x(w+\Delta w)-(w+\Delta w)^2+(w+\Delta w)^2+2 \Delta w x(w+\Delta w) \\
    =& 0,
    \end{aligned}
\end{equation}
which close the proof of the case when $frac(x)<\frac{\Delta w}{w+\Delta w} x+\frac{1}{2}$. Similar proof can be given for $frac(x)>\frac{\Delta w}{w+\Delta w} x+\frac{1}{2}$

\end{proof}

\section{More Experimental Setups}
\label{appendix:experimental_setup}

Our experiments on the ImageNet dataset and GLUE benchmark are based on the open-sourced code of ~\cite{wei2022qdrop} and ~\cite{peg}. Besides the settings mentioned in Sec. 4, we keep most of the settings the same as them. The first and the last layer are kept in 8-bit. The calibration set contains 1024 random images from ImageNet~\cite{5206848} and we finetune each block or layer for 20k iterations with batch size 32.  To maintain the gradient of parameters of the border function, we substitute the rounding operation by straight-through estimator (STE)~\cite{bengio2013estimating}. The learning rate for the activation step size is 4e-5 and for the weight rounding is 3e-3. Other different settings are mentioned below. We first introduce designs that necessary for optimizing activation rounding border function. Then we introduce settings that specific for ImageNet dataset and GLUE benchmark.

\subsection{Common Design of AQuant}
\label{appendix:implementation}
\begin{figure}[tbp]
  \centering
  \subfigure[Previous  pipeline]{\includegraphics[width=0.45\linewidth]{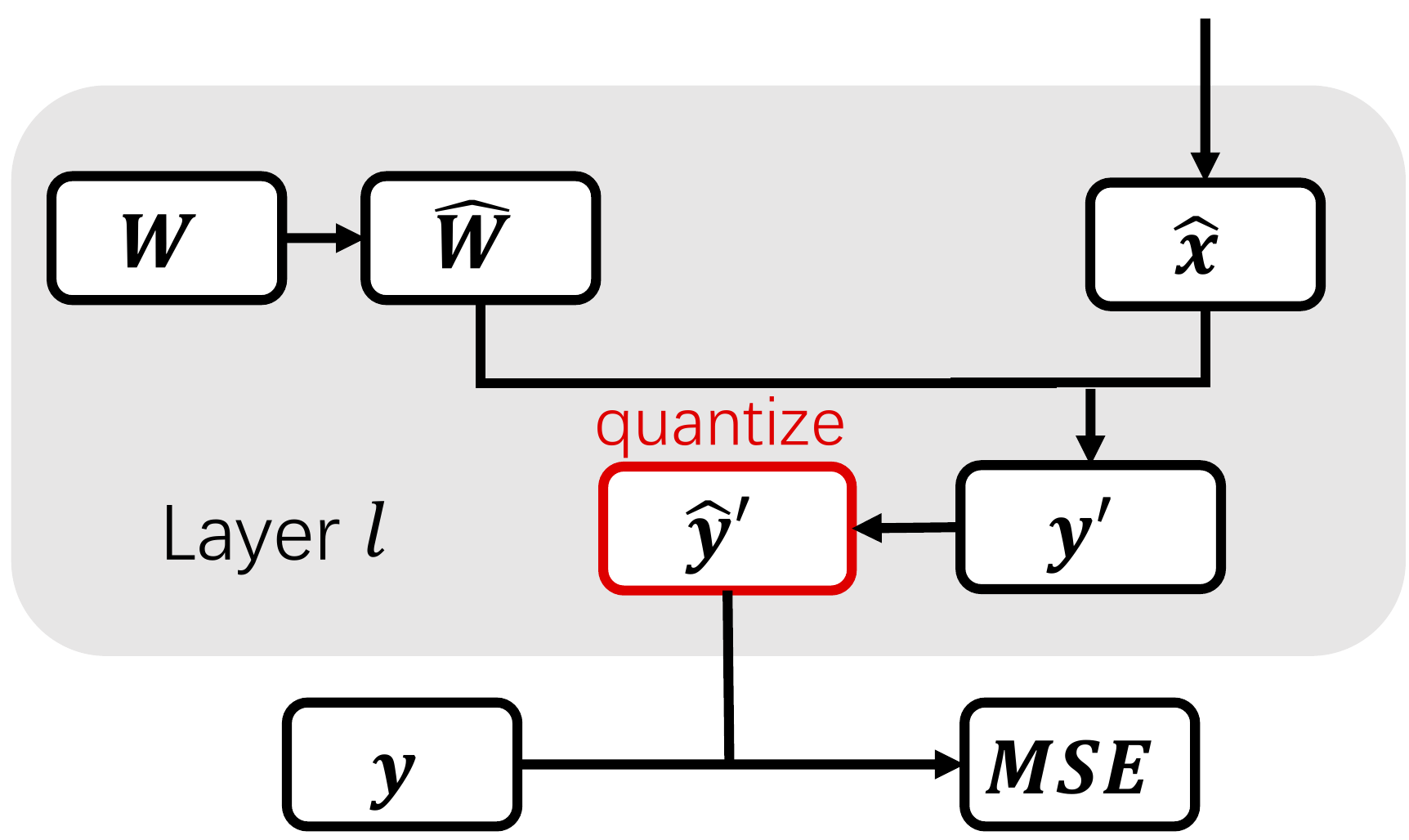} \label{fig:previous_quant}}
  \subfigure[AQuant  pipeline]{\includegraphics[width=0.45\linewidth]{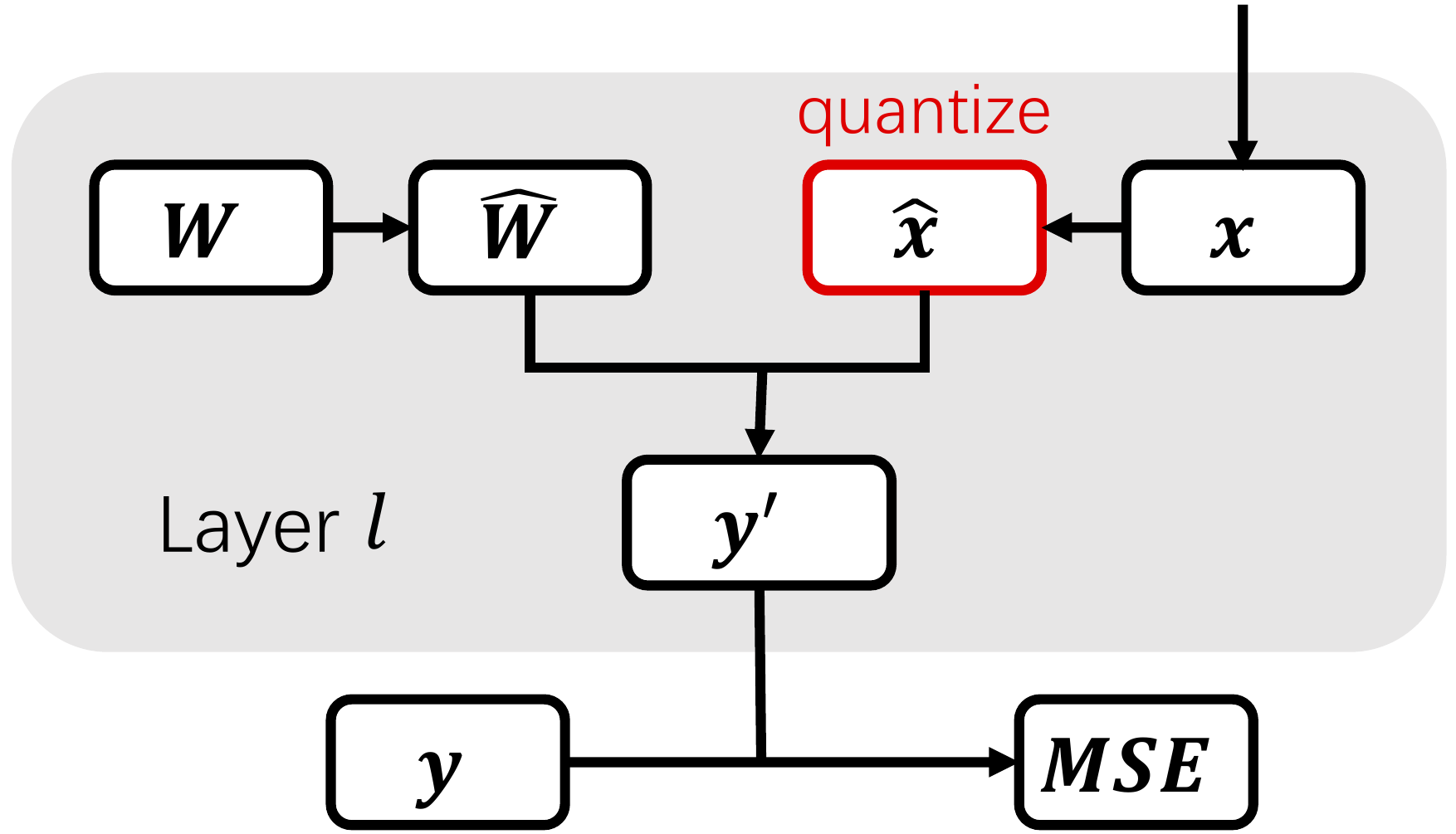} \label{fig:aquant_quant}}
  \caption{The position of activation quantization nodes of layer $l$. Positions of activation quantization are marked in a different color.}
\end{figure}

\paragraph{Refactor the position of activation quantization node} The quantization pipeline of previous works~\cite{nagel2020up,li2020brecq,wei2022qdrop} are shown in Figure \ref{fig:previous_quant}. The input of the layer $l$ is the activation that have been quantized at the previous layer. After the matrix multiplication, they apply the rounding-to-nearest scheme to outputs $\mathbf{y}^{\prime}$, and then compute the mean square error (MSE) between $\mathbf{\hat{y'}}$ and full-precision outputs $\mathbf{y}$. However, according to the optimization objective of AQuant, the rounding of activations $\mathbf{x}$ should depend on the output MSE of the layer $l$. Therefore, we refactor the quantization pipeline to the one in Figure \ref{fig:aquant_quant}. The layer $l$ accepts unquantized activations as the input, and then AQuant adaptively quantizes activations at the beginning of layer $l$. After the forward, we compute MSE between unquantized outputs and full-precision outputs.

One direct result of this refactor is that gradients of the activation quantization include the impact of weights, as shown in the following equation

\begin{equation}
\begin{aligned} 
\frac{\partial M S E}{\partial \mathbf{b}}=\frac{\partial M S E}{\partial \mathbf{y}^{\prime}} \frac{\partial \mathbf{y}^{\prime}}{\partial \mathbf{b}} &=\frac{\partial M S E}{\partial \mathbf{y}^{\prime}} \frac{\partial \widehat{\mathbf{W}} \widehat{\mathbf{x}}}{\partial \mathbf{b}} \\
&=\frac{\partial M S E}{\partial \mathbf{x}} \widehat{\mathbf{W}} \frac{\partial \hat{\mathbf{x}}}{\partial \mathbf{b}}
\end{aligned}
\end{equation}
where $\mathbf{b}$ are parameters related to the activation quantization, such as the border function $B(\cdot)$ or the learned scale step~\cite{esser2019learned}. 
\paragraph{Bound the border range} During the training, the border function is easily optimized to an ill status where the border value is substantially positive or negative. This crashes the optimization and never converges to a local minimum. To stable the training, we bound the value of the border function by the sigmoid function, whose output is bounded within $[0,1]$. In Equ. (6) of the original paper, the $\frac{\Delta w x}{w+\Delta w}$ may result in values greater than 0.5 or smaller than -0.5. Possible reasons include the large $\Delta$ by clipping, large activation value, and small weight value. Moreover, due to the computation nature of the sigmoid, its output is hard to approach 0 or 1. Therefore, we scale the output border value by a factor, which is 2.5 in our implementation. The sigmoid is also element-wise computed so that it can be fused with other operators with negligible overhead.

\paragraph{Rounding schedule}
During the finetuning process, the changeable border causes the same activation leaps among rounding up and rounding down. The jumping rounding error makes the training process unstable. Therefore, we let the rounding error of activations be gradually introduced. Consider an activation value ${x}$ and we let 
\begin{equation}
   \hat{x}={x}+ \alpha(\lceil {x}-B({x}) \rceil),
\end{equation}
where $\alpha$ controls the influence of rounding error. The $\alpha$ equals 1 represents normal quantization while the $\alpha$ equals 0 means no rounding error is introduced. AQuant sets $\alpha$ as 0 for the first 20\% finetuning iterations and then lets $\alpha$ linearly approach 1 as the finetuning goes on. We choose the warmup for the first  20\% iterations since the weight rounding scheme of the used baselines is also freedom optimized for the 20\% iterations.

\subsection{ImageNet Dataset}
\paragraph{Convergence of the weight rounding scheme}Previous works use soft quantization variables $h(\mathbf{V})$ to learn the rounding scheme of weights
\begin{equation}
\mathbf{W}'=\mathrm{s} \cdot \operatorname{clip}\left(\left\lfloor\frac{\mathbf{W}}{\mathrm{s}}\right\rfloor+h(\mathbf{V}), min, max\right).
\end{equation}
The function $h(\cdot)$ is a rectified sigmoid proposed in \cite{louizos2017learning} and $V$ is regularized by 
\begin{equation}
    f_{r e g}(\mathbf{V})=\lambda(\sum_{i, j} 1-\left|2 h\left(\mathbf{V}_{i, j}\right)-1\right|^{\beta}),
\end{equation}
where $\lambda$ is the harmony coefficient balancing the regularization and the output mean square error (MSE).
They anneal the parameter $\beta$ to gradually learn the rounding scheme. A higher $\beta$ allows $h(\mathbf{V})$ to freely adapt to improve the output MSE. As the finetuning process goes on, $\beta$ gradually decreases to encourage $h(\mathbf{V})$ to converge to 0 or 1, which finally arrives at the binary solution.  Specifically, the $\beta$ begins at 20 and finally decreases to 2 and the $\lambda$ is set as 0.01 in their works.  Through experiments, we find importing the border function slows down the convergence of $h(\mathbf{V})$ due to the flipping of activation rounding.  Since we keep the same number of finetuning iterations, for these layers, we let the $\beta$ stars with 16 and set $\lambda$ as 0.05 to enhance the regularization of weight rounding.

\paragraph{Co-optimization of the step size and weight rounding} 
\label{subsec:imagenet}
AdaRound~\cite{nagel2020up} and BRECQ~\cite{li2020brecq} sequentially optimize the weight rounding and the activation scale step. As our approach necessitates the concurrent training of the border function and weight rounding schemes, the step size is also jointly optimized during the finetuning stage. We observe that this joint optimization outperforms separate optimization. Recent studies, such as QDrop~\cite{wei2022qdrop}, also present this finding. To ensure a fair comparison, we adopt the joint optimization strategy for all methods in our experiments. Consequently, the gap between QDrop and the preceding methods is reduced.

\paragraph{The substitution of QDrop} Our implementation is based on the open-sourced code of QDrop~\cite{wei2022qdrop}. The main idea of QDrop is to find a flat minimum by randomly dropping the quantization of activations during finetuning. Dropping quantization means they substitute the noised activations with full-precision activations. QDrop finetunes the pre-trained model on the block level. Two types of dropping contribute to their improvements. The first is to drop the quantization of the block input and the second is to drop the quantization of intermediate activations within the block. In AQuant, only the dropping of block input is adopted. This is because dropping the quantization of intermediate activations changes the activation distribution and may mislead the optimization of the border function.

\subsection{GLUE Benchmark}

\paragraph{Co-optimization of the step size and weight rounding}The experiments of the GLUE benchmark are evaluated on the transformer model BERT. Our implementation on the transformer is based on the open-sourced code of ~\cite{peg}. They faithfully implement the method of AdaRound, in which the activation step size is post-optimized after the weight rounding. Due to the same reason in Sec. \ref{subsec:imagenet}, we optimize the activation step size together with the weight rounding. Therefore, the reported results of the AdaRound are much stronger than the original paper.

\section{Border Function+}

We show the linear border function is a feasible solution to adaptively adjust the activation rounding with almost negligible overhead. This section further introduces another two factors that affect the {\verror} and our attempts to include them in the border function. Note that any improvement should be subject to the overhead.

\subsection{Two attempts}
\label{subsec:more_consideration}
\begin{figure}[tbp] 
    \centering 
    \includegraphics[width=0.89\linewidth]{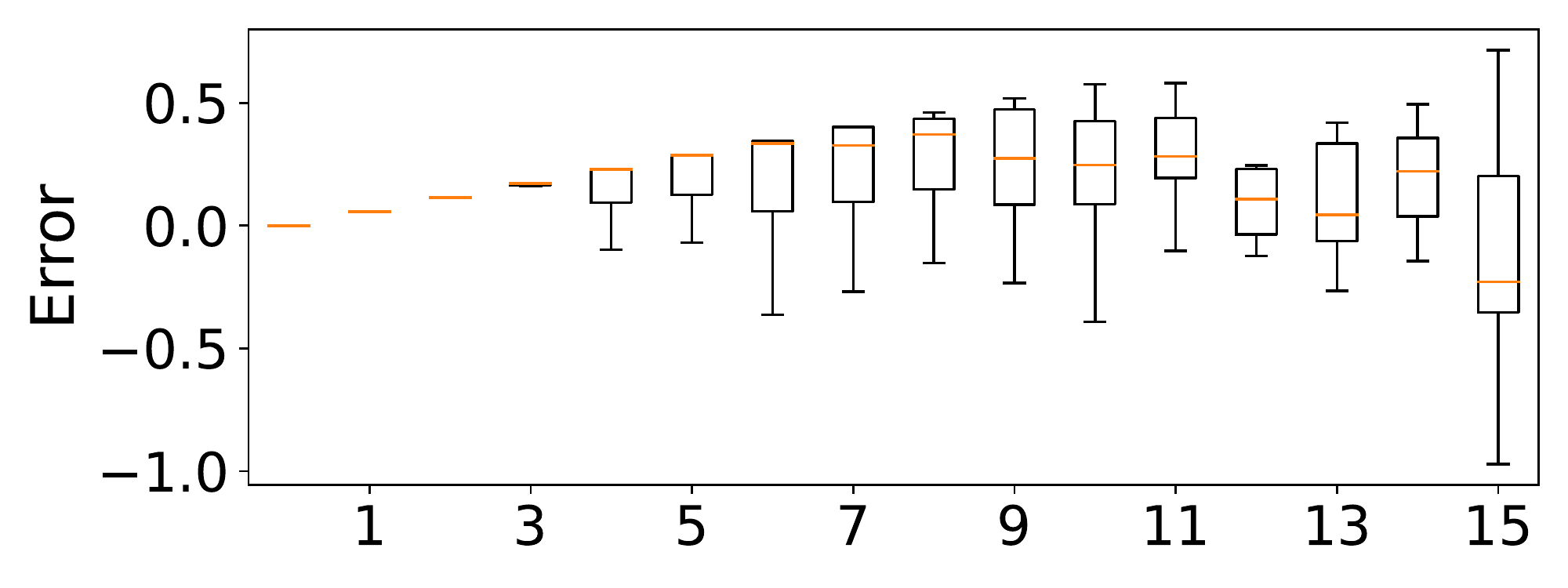}  
    \caption{The propagated error w.r.t. the noised activation $x'$. We group $x'$ into 16 clusters according to their magnitude, as listed on the X-axis. Y-axis is the error between noised and oracle activation. } 
    \label{fig:e_wrt_x}
\end{figure}

\paragraph{Propagated Error} The intermediate activations include the propagated errors caused by previous layers. Prior works~\cite{nagel2020up,li2020brecq,wei2022qdrop} handle it implicitly by end-to-end optimization. While we need to verify whether our border function can capture the impact of the propagated error so we consider it explicitly. To this end, we further include the propagated error in the objective Equ. (4) of the original paper
\begin{equation}
\begin{split}
& \mathop{\arg\min}\limits_{\Delta \mathbf{x}, \Delta \mathbf{W}} \mathbb{E}\left[\left(\left(\mathbf{W}_{i, :}+\Delta \mathbf{W}_{i, :}\right)\left(\mathbf{x'}+\Delta \mathbf{x'}\right)-\mathbf{W}_{i ,:} \mathbf{x}\right)^{2}\right]. 
\label{equ:objective_with_activation_propagatederror}
\end{split}
\end{equation}
Note we change the input from $\mathbf{x}$ to the noised version $\mathbf{x'}$, and the $\Delta \mathbf{x'}$ is the quantization error based on the noised input. Following the same way in Sec. \ref{appendix:derivation}, the rounding border becomes
\begin{equation}
B^E(x')=\frac{\Delta w}{w+\Delta w}x' + \frac{ w}{w+\Delta w}e(x') + \frac{1}{2},
\label{equ:linear_border_withPropagatedError}
\end{equation}
where $e(x')=x'-x$ is the propagated error. The newly introduced term $\frac{w}{w+\Delta w}e(x')$ defines the impact of the propagated error. However, the propagated error $e(x')$ is determined by the quantization of previous layers, which is a complex process and is hard to analyze. Therefore, we present an empirical study to handle it through experiments. Our experiments are conducted on ResNet-18 under W2A4 quantization. For an element of the second layer input, we summarize activation values and their errors on 1024 images. Since the complex process in previous layers produces the errors in a random-like fashion, we only consider fitting the propagated error in the sense of its mean value. As shown in Fig. \ref{fig:e_wrt_x}, when the magnitude of $x'$ increases, the mean value first slowly deviates from zero and then moves towards zero again. The non-linear trends are also observed on other pixels and other layers.
In order for the border function to be able to capture this nonlinear relationship, we propose to extend Equ. (6) of the original paper to a quadratic function w.r.t. $x'$
\begin{equation}
B^E(x')=b_{2} \cdot x'^2 + b_{1} \cdot x'+b_{0}.
\label{equ:quadratic_border}
\end{equation}
$b_{:}$ are parameters learned on the calibration set. 
Quadratic function approximately introduces the impact of the propagated error with trivial extra overhead on top of the linear border.

\paragraph{Border Fusion} The derived linear (quadratic if considering the propagated error) border is only aware of local information. As the border function is optimized using the {\verror} as the objective, a border that integrates global information is supposed to be a better solution. 
As such, we further consider optimizing the border function on a larger scope. To achieve this, we simply take an average for borders to integrate information. Consider the convolutional layer, the activation vector is composed of $i_c$ input channels, with each containing $k^2$ elements. The intuitive way is to average the border on the whole vector. However, activations and weights from different input channels have varying magnitudes. The gradient descent may only focus on input channels with large scales, which could result in a bad local minimum. Therefore, we set the scope as an input channel and the border is
\begin{equation}
B_i^{C_i}\left(\mathbf{x'}_{ik^2:(i+1)k^2}\right)=  \frac{ 1}{k^2} \sum_{j=ik^2}^{(i+1)k^2-1}  B^E_j(x'_j),
\label{equ:final_B(x)}
\end{equation}
where $\mathbf{x'}_{ik^2:(i+1)k^2}$ is the $i_{th}$ input channel of vector $\mathbf{x'}$ and $B^E_j$ is the border function of the $j_{th}$ element $x_j$. The superscript $C_{i}$ indicate the border is averaged within an input channel. The border value is averaged and then shared within the input channel. This method may not be the optimal way to integrate global information, but it introduces no additional parameters, and the introduced average operation can still be easily fused with the \texttt{img2col} kernel.

\subsection{Ablation Study}

\begin{table}[tbp]
\footnotesize
\centering
    
  \caption{Ablation experiments of border function and fusion.}
    \begin{tabular}{ccccc}
    \toprule
    Methods & Strategy & Bits & Res18 & Reg600M \\
    \midrule
    \multicolumn{1}{c}{\multirow{4}[4]{*}{\makecell[c]{Border \\ order}}} & Linear & W2A2 & 59.14 & 45.37 \\
       & Quadratic & W2A2 & \textbf{60.31} & \textbf{46.61} \\
    \cmidrule{2-5}   & Linear & W3A3 & 67.59 & 66.78 \\
       & Quadratic & W3A3 & \textbf{67.97} & \textbf{67.11} \\
    \midrule
    \multicolumn{1}{c}{\multirow{4}[4]{*}{\makecell[c]{Border \\ fusion}}} & No fusion & W2A2 & 58.56 & 44.95 \\
       & Fusion & W2A2 & \textbf{60.31} & \textbf{46.61} \\
    \cmidrule{2-5}   & No fusion & W3A3 & 67.76 & 66.93 \\
       & Fusion & W3A3 & \textbf{67.97} & \textbf{67.11} \\
    \bottomrule
    \end{tabular}%

  \label{tab:ablation}%
\end{table}%

\label{subsec:ablation_exp}
This section gives ablation of two techniques mentioned in Sec. \ref{subsec:more_consideration}.
\paragraph{Border Order} 
We evaluate the efficacy of the quadratic border in reducing the propagated error compared with the linear border. Other settings are kept the same. In Table \ref{tab:ablation}, the quadratic border outperforms the linear border on both models and bitwidths.

\paragraph{Border Fusion}
We then test the effectiveness of border fusion on input channels. ``Fusion'' is exact the AQuant mentioned in Sec. 4 and ``No fusion'' uses the element-wise border function. According to Table \ref{tab:ablation}, border fusion further improves the accuracy. On extremely low bitwidth, fusion produces more apparent improvements.

We find that the gap of both border function and border fusion shrink for 3-bit quantization because the effects of propagated errors and rounding errors get small. 
This also gives the freedom to use the linear border function or the element-wise function for high-bit quantization to save more overhead.

%% file: anonymous-submission-latex-2024.bbl
\begin{thebibliography}{35}
\providecommand{\natexlab}[1]{#1}

\bibitem[{Banner et~al.(2019)Banner, Nahshan, Hoffer, and
  Soudry}]{banner2018aciq}
Banner, R.; Nahshan, Y.; Hoffer, E.; and Soudry, D. 2019.
\newblock {ACIQ}: Analytical Clipping for Integer Quantization of neural
  networks.

\bibitem[{Banner, Nahshan, and Soudry(2019)}]{banner2019post}
Banner, R.; Nahshan, Y.; and Soudry, D. 2019.
\newblock Post training 4-bit quantization of convolutional networks for
  rapid-deployment.
\newblock \emph{Advances in Neural Information Processing Systems}, 32.

\bibitem[{Bondarenko, Nagel, and Blankevoort(2021)}]{peg}
Bondarenko, Y.; Nagel, M.; and Blankevoort, T. 2021.
\newblock Understanding and Overcoming the Challenges of Efficient Transformer
  Quantization.
\newblock In \emph{Proceedings of the 2021 Conference on Empirical Methods in
  Natural Language Processing}, 7947--7969. Online and Punta Cana, Dominican
  Republic: Association for Computational Linguistics.

\bibitem[{Cai et~al.(2020)Cai, Yao, Dong, Gholami, Mahoney, and
  Keutzer}]{cai2020zeroq}
Cai, Y.; Yao, Z.; Dong, Z.; Gholami, A.; Mahoney, M.~W.; and Keutzer, K. 2020.
\newblock Zeroq: A novel zero shot quantization framework.
\newblock In \emph{Proceedings of the IEEE/CVF Conference on Computer Vision
  and Pattern Recognition}, 13169--13178.

\bibitem[{Choi et~al.(2018)Choi, Wang, Venkataramani, Chuang, Srinivasan, and
  Gopalakrishnan}]{choi2018pact}
Choi, J.; Wang, Z.; Venkataramani, S.; Chuang, P. I.-J.; Srinivasan, V.; and
  Gopalakrishnan, K. 2018.
\newblock Pact: Parameterized clipping activation for quantized neural
  networks.
\newblock \emph{arXiv preprint arXiv:1805.06085}.

\bibitem[{Choi et~al.(2021)Choi, Hong, Park, Kim, and
  Lee}]{NEURIPS2021_7cc23420}
Choi, K.; Hong, D.; Park, N.; Kim, Y.; and Lee, J. 2021.
\newblock Qimera: Data-free Quantization with Synthetic Boundary Supporting
  Samples.
\newblock In Ranzato, M.; Beygelzimer, A.; Dauphin, Y.; Liang, P.; and Vaughan,
  J.~W., eds., \emph{Advances in Neural Information Processing Systems},
  volume~34, 14835--14847. Curran Associates, Inc.

\bibitem[{Choukroun et~al.(2019)Choukroun, Kravchik, Yang, and
  Kisilev}]{choukroun2019low}
Choukroun, Y.; Kravchik, E.; Yang, F.; and Kisilev, P. 2019.
\newblock Low-bit quantization of neural networks for efficient inference.
\newblock In \emph{2019 IEEE/CVF International Conference on Computer Vision
  Workshop (ICCVW)}, 3009--3018. IEEE.

\bibitem[{Deng et~al.(2009)Deng, Dong, Socher, Li, Li, and Fei-Fei}]{5206848}
Deng, J.; Dong, W.; Socher, R.; Li, L.-J.; Li, K.; and Fei-Fei, L. 2009.
\newblock ImageNet: A large-scale hierarchical image database.
\newblock In \emph{2009 IEEE Conference on Computer Vision and Pattern
  Recognition}, 248--255.

\bibitem[{Esser et~al.(2019)Esser, McKinstry, Bablani, Appuswamy, and
  Modha}]{esser2019learned}
Esser, S.~K.; McKinstry, J.~L.; Bablani, D.; Appuswamy, R.; and Modha, D.~S.
  2019.
\newblock LEARNED STEP SIZE QUANTIZATION.
\newblock In \emph{International Conference on Learning Representations}.

\bibitem[{Finkelstein, Almog, and
  Grobman(2019{\natexlab{a}})}]{finkelstein2019fighting}
Finkelstein, A.; Almog, U.; and Grobman, M. 2019{\natexlab{a}}.
\newblock Fighting quantization bias with bias.
\newblock \emph{arXiv preprint arXiv:1906.03193}.

\bibitem[{Finkelstein, Almog, and
  Grobman(2019{\natexlab{b}})}]{DBLP:journals/corr/abs-1906-03193}
Finkelstein, A.; Almog, U.; and Grobman, M. 2019{\natexlab{b}}.
\newblock Fighting Quantization Bias With Bias.
\newblock \emph{CoRR}, abs/1906.03193.

\bibitem[{Frantar et~al.(2023)Frantar, Ashkboos, Hoefler, and
  Alistarh}]{frantar2023optq}
Frantar, E.; Ashkboos, S.; Hoefler, T.; and Alistarh, D. 2023.
\newblock {OPTQ}: Accurate Quantization for Generative Pre-trained
  Transformers.
\newblock In \emph{The Eleventh International Conference on Learning
  Representations}.

\bibitem[{Gong et~al.(2019)Gong, Liu, Jiang, Li, Hu, Lin, Yu, and
  Yan}]{gong2019differentiable}
Gong, R.; Liu, X.; Jiang, S.; Li, T.; Hu, P.; Lin, J.; Yu, F.; and Yan, J.
  2019.
\newblock Differentiable soft quantization: Bridging full-precision and low-bit
  neural networks.
\newblock In \emph{Proceedings of the IEEE/CVF International Conference on
  Computer Vision}, 4852--4861.

\bibitem[{Guo et~al.(2021)Guo, Qiu, Leng, Gao, Zhang, Liu, Yang, Zhu, and
  Guo}]{guo2021squant}
Guo, C.; Qiu, Y.; Leng, J.; Gao, X.; Zhang, C.; Liu, Y.; Yang, F.; Zhu, Y.; and
  Guo, M. 2021.
\newblock SQuant: On-the-Fly Data-Free Quantization via Diagonal Hessian
  Approximation.
\newblock In \emph{International Conference on Learning Representations}.

\bibitem[{He et~al.(2016)He, Zhang, Ren, and Sun}]{he2016deep}
He, K.; Zhang, X.; Ren, S.; and Sun, J. 2016.
\newblock Deep residual learning for image recognition.
\newblock In \emph{Proceedings of the IEEE conference on computer vision and
  pattern recognition}, 770--778.

\bibitem[{Horn and Johnson(1985)}]{horn_johnson_1985}
Horn, R.~A.; and Johnson, C.~R. 1985.
\newblock \emph{Matrix Analysis}.
\newblock Cambridge University Press.

\bibitem[{Hubara et~al.(2021)Hubara, Nahshan, Hanani, Banner, and
  Soudry}]{hubara2021accurate}
Hubara, I.; Nahshan, Y.; Hanani, Y.; Banner, R.; and Soudry, D. 2021.
\newblock Accurate post training quantization with small calibration sets.
\newblock In \emph{International Conference on Machine Learning}, 4466--4475.
  PMLR.

\bibitem[{Jacob et~al.(2018)Jacob, Kligys, Chen, Zhu, Tang, Howard, Adam, and
  Kalenichenko}]{jacob2018quantization}
Jacob, B.; Kligys, S.; Chen, B.; Zhu, M.; Tang, M.; Howard, A.; Adam, H.; and
  Kalenichenko, D. 2018.
\newblock Quantization and training of neural networks for efficient
  integer-arithmetic-only inference.
\newblock In \emph{Proceedings of the IEEE conference on computer vision and
  pattern recognition}, 2704--2713.

\bibitem[{Jia et~al.(2014)Jia, Shelhamer, Donahue, Karayev, Long, Girshick,
  Guadarrama, and Darrell}]{jia2014caffe}
Jia, Y.; Shelhamer, E.; Donahue, J.; Karayev, S.; Long, J.; Girshick, R.;
  Guadarrama, S.; and Darrell, T. 2014.
\newblock Caffe: Convolutional architecture for fast feature embedding.
\newblock In \emph{Proceedings of the 22nd ACM international conference on
  Multimedia}, 675--678.

\bibitem[{Li et~al.(2021)Li, Gong, Tan, Yang, Hu, Zhang, Yu, Wang, and
  Gu}]{li2020brecq}
Li, Y.; Gong, R.; Tan, X.; Yang, Y.; Hu, P.; Zhang, Q.; Yu, F.; Wang, W.; and
  Gu, S. 2021.
\newblock BRECQ: Pushing the Limit of Post-Training Quantization by Block
  Reconstruction.
\newblock In \emph{International Conference on Learning Representations}.

\bibitem[{Liu et~al.(2021)Liu, Wang, Han, Zhang, Ma, and
  Gao}]{NEURIPS2021_ec895663}
Liu, Z.; Wang, Y.; Han, K.; Zhang, W.; Ma, S.; and Gao, W. 2021.
\newblock Post-Training Quantization for Vision Transformer.
\newblock In Ranzato, M.; Beygelzimer, A.; Dauphin, Y.; Liang, P.; and Vaughan,
  J.~W., eds., \emph{Advances in Neural Information Processing Systems},
  volume~34, 28092--28103. Curran Associates, Inc.

\bibitem[{Nagel et~al.(2020)Nagel, Amjad, Van~Baalen, Louizos, and
  Blankevoort}]{nagel2020up}
Nagel, M.; Amjad, R.~A.; Van~Baalen, M.; Louizos, C.; and Blankevoort, T. 2020.
\newblock Up or down? adaptive rounding for post-training quantization.
\newblock In \emph{International Conference on Machine Learning}, 7197--7206.
  PMLR.

\bibitem[{Nagel et~al.(2019)Nagel, Baalen, Blankevoort, and
  Welling}]{nagel2019data}
Nagel, M.; Baalen, M.~v.; Blankevoort, T.; and Welling, M. 2019.
\newblock Data-free quantization through weight equalization and bias
  correction.
\newblock In \emph{Proceedings of the IEEE/CVF International Conference on
  Computer Vision}, 1325--1334.

\bibitem[{Niu et~al.(2021)Niu, Guan, Wang, Agrawal, and Ren}]{niu2021dnnfusion}
Niu, W.; Guan, J.; Wang, Y.; Agrawal, G.; and Ren, B. 2021.
\newblock DNNFusion: accelerating deep neural networks execution with advanced
  operator fusion.
\newblock In \emph{Proceedings of the 42nd ACM SIGPLAN International Conference
  on Programming Language Design and Implementation}, 883--898.

\bibitem[{Paszke et~al.(2019)Paszke, Gross, Massa, Lerer, Bradbury, Chanan,
  Killeen, Lin, Gimelshein, Antiga et~al.}]{paszke2019pytorch}
Paszke, A.; Gross, S.; Massa, F.; Lerer, A.; Bradbury, J.; Chanan, G.; Killeen,
  T.; Lin, Z.; Gimelshein, N.; Antiga, L.; et~al. 2019.
\newblock Pytorch: An imperative style, high-performance deep learning library.
\newblock \emph{Advances in neural information processing systems}, 32.

\bibitem[{Radosavovic et~al.(2020)Radosavovic, Kosaraju, Girshick, He, and
  Doll{\'a}r}]{radosavovic2020designing}
Radosavovic, I.; Kosaraju, R.~P.; Girshick, R.; He, K.; and Doll{\'a}r, P.
  2020.
\newblock Designing network design spaces.
\newblock In \emph{Proceedings of the IEEE/CVF Conference on Computer Vision
  and Pattern Recognition}, 10428--10436.

\bibitem[{Sandler et~al.(2018)Sandler, Howard, Zhu, Zhmoginov, and
  Chen}]{sandler2018mobilenetv2}
Sandler, M.; Howard, A.; Zhu, M.; Zhmoginov, A.; and Chen, L.-C. 2018.
\newblock Mobilenetv2: Inverted residuals and linear bottlenecks.
\newblock In \emph{Proceedings of the IEEE conference on computer vision and
  pattern recognition}, 4510--4520.

\bibitem[{Tan et~al.(2019)Tan, Chen, Pang, Vasudevan, Sandler, Howard, and
  Le}]{tan2019mnasnet}
Tan, M.; Chen, B.; Pang, R.; Vasudevan, V.; Sandler, M.; Howard, A.; and Le,
  Q.~V. 2019.
\newblock Mnasnet: Platform-aware neural architecture search for mobile.
\newblock In \emph{Proceedings of the IEEE/CVF Conference on Computer Vision
  and Pattern Recognition}, 2820--2828.

\bibitem[{Vaswani et~al.(2017)Vaswani, Shazeer, Parmar, Uszkoreit, Jones,
  Gomez, Kaiser, and Polosukhin}]{vaswani2017attention}
Vaswani, A.; Shazeer, N.; Parmar, N.; Uszkoreit, J.; Jones, L.; Gomez, A.~N.;
  Kaiser, {\L}.; and Polosukhin, I. 2017.
\newblock Attention is all you need.
\newblock \emph{Advances in neural information processing systems}, 30.

\bibitem[{Wang et~al.(2020)Wang, Chen, He, and Cheng}]{wang2020towards}
Wang, P.; Chen, Q.; He, X.; and Cheng, J. 2020.
\newblock Towards accurate post-training network quantization via bit-split and
  stitching.
\newblock In \emph{International Conference on Machine Learning}, 9847--9856.
  PMLR.

\bibitem[{Wang et~al.(2019)Wang, Lu, Tao, Zhou, and Tian}]{wang2019learning}
Wang, Z.; Lu, J.; Tao, C.; Zhou, J.; and Tian, Q. 2019.
\newblock Learning channel-wise interactions for binary convolutional neural
  networks.
\newblock In \emph{Proceedings of the IEEE/CVF Conference on Computer Vision
  and Pattern Recognition}, 568--577.

\bibitem[{Wei et~al.(2022)Wei, Gong, Li, Liu, and Yu}]{wei2022qdrop}
Wei, X.; Gong, R.; Li, Y.; Liu, X.; and Yu, F. 2022.
\newblock {QD}rop: Randomly Dropping Quantization for Extremely Low-bit
  Post-Training Quantization.
\newblock In \emph{International Conference on Learning Representations}.

\bibitem[{Wolf et~al.(2020)Wolf, Debut, Sanh, Chaumond, Delangue, Moi, Cistac,
  Rault, Louf, Funtowicz, Davison, Shleifer, von Platen, Ma, Jernite, Plu, Xu,
  Scao, Gugger, Drame, Lhoest, and Rush}]{huggingface}
Wolf, T.; Debut, L.; Sanh, V.; Chaumond, J.; Delangue, C.; Moi, A.; Cistac, P.;
  Rault, T.; Louf, R.; Funtowicz, M.; Davison, J.; Shleifer, S.; von Platen,
  P.; Ma, C.; Jernite, Y.; Plu, J.; Xu, C.; Scao, T.~L.; Gugger, S.; Drame, M.;
  Lhoest, Q.; and Rush, A.~M. 2020.
\newblock Transformers: State-of-the-Art Natural Language Processing.
\newblock In \emph{Proceedings of the 2020 Conference on Empirical Methods in
  Natural Language Processing: System Demonstrations}, 38--45. Online:
  Association for Computational Linguistics.

\bibitem[{Xiao et~al.(2023)Xiao, Lin, Seznec, Wu, Demouth, and
  Han}]{xiao2023smoothquant}
Xiao, G.; Lin, J.; Seznec, M.; Wu, H.; Demouth, J.; and Han, S. 2023.
\newblock Smoothquant: Accurate and efficient post-training quantization for
  large language models.
\newblock In \emph{International Conference on Machine Learning}, 38087--38099.
  PMLR.

\bibitem[{Zhao et~al.(2019)Zhao, Hu, Dotzel, De~Sa, and
  Zhang}]{zhao2019improving}
Zhao, R.; Hu, Y.; Dotzel, J.; De~Sa, C.; and Zhang, Z. 2019.
\newblock Improving neural network quantization without retraining using
  outlier channel splitting.
\newblock In \emph{International conference on machine learning}, 7543--7552.
  PMLR.

\end{thebibliography}
